\documentclass[10pt, conference]{IEEEtran} %
\usepackage{algorithm}
\usepackage{algpseudocode}
\usepackage{cite} 
\usepackage{amsmath,amssymb,amsfonts}
\usepackage{graphicx}
\usepackage{booktabs}
\usepackage{url}

\begin{document}

\title{Workload-Aware Caching for Multi-Agent Systems}

\author{
\IEEEauthorblockN{Anas Mohamed$^1$,
Kaizan Haque$^1$,
Azal Ahmad Khan$^1$,
Chetan Sharma$^2$,
Shuwen Ge$^3$,
Ali Anwar$^1$}
\IEEEauthorblockA{$^1$University of Minnesota \quad $^2$Google \quad $^3$IIT Guwahati}
}

\maketitle

\begin{abstract}
Multi-agent systems decompose complex tasks into directed acyclic graphs (DAGs) of specialized agent executions, creating natural opportunities for caching intermediate results across queries. However, existing cache eviction policies treat all cached entries uniformly based on access history, ignoring structural and workload signals uniquely available in agentic execution environments. We present a workload-aware eviction policy that combines three signals, namely recomputation cost, DAG dependency count, and agent invocation frequency, into a unified scoring function that retains the most valuable entries under memory constraints. Evaluated across three multi-agent benchmarks spanning diverse reuse regimes, our policy reduces latency by up to 64.7\% relative to the uncached baseline and achieves on average a 31.1\% latency reduction over the next best finite-capacity baseline, while approaching the performance of an unbounded cache and maintaining accuracy on par with or exceeding all competing finite-capacity methods. We further show that workload-aware content caching is complementary to other agentic system optimization methods, including plan-level caching and parallel agent execution, with each technique targeting a distinct efficiency bottleneck in multi-agent pipelines.
\end{abstract}

\begin{IEEEkeywords}
Multi-agent systems, cache eviction, directed acyclic graphs, workload-aware caching
\end{IEEEkeywords}

\section{Introduction}
AI systems are moving from single model architectures towards compound systems that use and coordinate different specialized agents, tools, and reasoning steps to solve queries. Multi agent systems (MAS) have shown capabilities across domains from software engineering to scientific research to autonomous decision making~\cite{jimenez2024swebench,chen2025aiopslab,lu2024aiscientist,yamada2025aiscientistv2,yao2023react,Wang2023ASO}. However, this architectural shift has also introduced efficiency challenges, and as agent interactions become more complex with more elaborate plans, computational costs can limit practical deployment in the real world. The inference costs of modern reasoning models combined with the need for multiple agent invocations per query makes every redundant computation quite expensive~\cite{liu2025efficient}, not just in latency but in resource utilization as well. 

Multi agent workloads can show substantial computational overlap across queries. This redundancy comes up when different queries share common computational patterns. To give an example, in financial analysis, queries like "What was Company X’s revenue?" and "What was Company X’s profit margin?" both share identical initial steps like retrieving financial statements, parsing documents, and extracting standard fields, differing only in the final calculations. Similarly, document analysis queries asking for summaries versus dataset descriptions might share the same parsing, section identification, and entity extraction operations. The pattern extends beyond exact repetition as well. Structurally similar queries like "Analyze Q1 2023 performance" and "Analyze Q2 2023 performance" might differ only in date filters but share data aggregation, normalization, and visualization logic.

\begin{figure}[t]
\centering
\includegraphics[width=0.65\columnwidth]{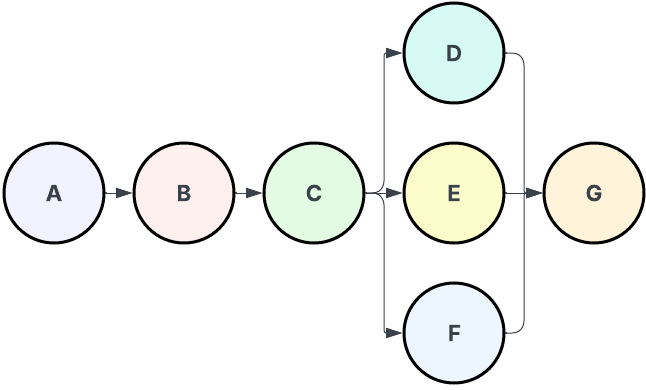}
\caption{Multi-agent workflow for network monitoring represented as a DAG. Nodes represent steps completed by specialized agents with edges encoding dependencies. A) Telemetry Collection, B) Data Parsing and Normalization, C) Metric Aggregation and Calculation, D) Capacity/Usage Analysis, E) Anomaly Detection, F) SLA Policy Evaluation, G) Dashboard and Alert Updates. Expensive operations benefit significantly from caching.}
\label{fig:dag-example}
\end{figure}

Many modern multi agent systems use DAG-based planning where specialized planning agents decompose tasks into structured execution plans (Figure~\ref{fig:dag-example}). These plans are represented as directed acyclic graphs where nodes correspond to computational tasks (LLM calls, tool executions, data processing) and edges encode dependencies between those tasks. An executor then dispatches agents to complete the task while respecting dependencies. This decomposition into reusable computational units leads to a natural caching opportunity, where if we cache node results, we can reuse them when similar queries require the same operations. However, unbounded caching quickly finishes available memory, which necessitates eviction policies that decide which cached results to retain.

Traditional eviction strategies like Least Recently Used (LRU) or Least Frequently Used (LFU) make decisions based purely on access history, with metrics like recency or frequency, treating all cached items uniformly. This approach is functional, but it ignores several structural and workload signals that are available in multi agent planning systems. For instance, consider a cache at capacity containing two nodes. Node A was accessed 30 seconds ago, feeds into 1 downstream calculation (so it has one outgoing edge), and takes 0.3 seconds to recompute. Node B was accessed 5 minutes ago, feeds into 4 downstream nodes, and takes 8 seconds to recompute due to expensive database queries and document parsing. LRU evicts Node B based on older access time. When the next query arrives requiring Node B's outputs but not Node A’s, the system must recompute it (8 seconds). Evicting Node A would have cost only 0.3 seconds. This is just one example of uniform treatment that fails to recognize that Node B's structural importance (feeding multiple dependents) and high recomputation cost make it more valuable to retain. Beyond structural considerations, traditional policies also do not exploit workload patterns. When users explore related questions (e.g. analyzing different metrics for the same company, or testing code variations against the same test suite), certain DAG subgraphs become temporarily more valuable based on query similarity. A policy aware of these semantic patterns could prioritize retaining nodes likely to be reused in future queries.

This paper introduces a workload aware, structure informed cache eviction policy that is specifically designed for multi agent systems. Our approach considers three dimensions that most traditional policies ignore. (1) Topological importance of nodes within DAG structures, (2) computational cost of recomputing evicted nodes, and (3) likelihood of future reuse based on observed workload patterns. By combining these signals into a unified scoring function, our policy reduces latency by up to 64.7\% relative to the uncached baseline, while approaching the performance of an unbounded cache across diverse multi-agent workloads. We also analyze how our approach interacts with complementary multi-agent efficiency techniques, finding that workload-aware content caching, plan-level caching, and parallel agent execution are mutually reinforcing. Overall, our contributions are:

\begin{itemize}
\item[\textbf{C1}] \textbf{Workload-Aware Eviction Policy.} A cache eviction policy for multi-agent systems that combines recomputation cost, DAG dependency count, and agent invocation frequency into a unified scoring function, enabling principled eviction decisions. 

\item[\textbf{C2}] \textbf{Empirical Evaluation.} A comprehensive evaluation across three multi-agent benchmarks spanning different reuse regimes, demonstrating consistent improvements in latency, throughput, and accuracy relative to standard eviction baselines. 

\item[\textbf{C3}] \textbf{Eviction Quality Analysis.} An analysis showing that eviction quality, not just hit rate, is also a primary driver of latency gains in heterogeneous agentic workloads, with our policy consistently evicting cheaper entries than all competing methods. 

\item[\textbf{C4}] \textbf{Agentic Optimization Study.} An investigation of how workload-aware caching interacts with complementary techniques including plan-level caching and parallel agent execution, revealing synergies. 

\end{itemize}
\section{Background and Related Work}
We organize related work into three areas. Caching techniques for LLM based agents, workflow abstractions using DAGs, and efficient multi-agent system orchestration.
\subsection{Caching in LLM Agents}
As LLM-based agents have grown in complexity and deployment scale, caching has emerged as a major optimization technique at multiple different levels of abstraction. We organize approaches by the granularity at which they cache computational artifacts. 
\paragraph{KV Cache}
KV-caching refers to caching key-value tensors from transformer attention computations~\cite{vaswani2017attention}. Modern serving systems like vLLM~\cite{kwon2023efficient} and SGLang~\cite{zheng2024sglang} introduced paged storage and radix-tree organization for efficient KV-cache management, usually employing LRU-based eviction when GPU memory becomes constrained. Recent work extends this specifically for multi-agent systems. KVFlow~\cite{pan2025kvflowefficientprefixcaching} introduces workflow-aware eviction using an Agent Step Graph to predict execution order, assigning each agent a steps-to-execution value estimating temporal proximity to future activation and evicting caches of agents unlikely to run soon. KVCOMM~\cite{ye2025kvcomm} addresses a complementary challenge, enabling KV cache sharing across agents with different prefix contexts by maintaining an anchor pool storing observed KV-cache offset deviations under varying prefixes, enabling cache reuse.
\paragraph{Semantic and Tool-Level Caching}
Higher level systems cache LLM outputs based on query similarity rather than exact matches. GPTCache~\cite{bang-2023-gptcache} pioneered embedding-based semantic caching for LLM applications, using similarity search~\cite{johnson2019billion}  to reuse responses for semantically related queries. Building on this foundation, Cortex~\cite{ruan2026cortexachievinglowlatencycostefficient} extends semantic caching to tool outputs in agent workloads, addressing cross-region data access latency through semantic-aware knowledge caching. These methods, and similar ones, target individual LLM or tool invocations, reducing redundant computation when similar queries recur.
\paragraph{Plan and Reasoning-Level Caching}
The highest level targets planning and reasoning stages of agent execution. Agentic Plan Caching~\cite{zhang2025agentic} extracts structured plan templates from completed executions, storing generalized workflows that can be adapted to new queries with similar task intents through keyword-based matching and lightweight template adaptation. Approaches such as those seen in SemanticALLI~\cite{chillara2026semanticallicachingreasoningjust} complement this by caching intermediate reasoning representations at multiple checkpoints within analytics pipelines, treating structured reasoning states as first-class cacheable artifacts. These approaches operate at fundamentally different granularities. KV cache methods optimize inference latency, semantic caching targets individual calls, and plan level systems reduce planning overhead. Our work introduces caching at the level of task execution results.
\subsection{Workflows Represented as DAGs}
Directed acyclic graphs naturally express computational workflows with explicit dependencies, enabling dependency-aware optimizations across multiple domains.
\paragraph{Data Analytics Systems.}
DAG-aware caching emerged in analytics frameworks where batch jobs decompose into stages with data dependencies. LRC~\cite{8057007} introduced reference-count-based eviction for Apache Spark~\cite{zaharia2012resilient}, prioritizing cached data blocks with the highest number of uncomputed children (an indicator of future utility that outperforms recency) or frequency-based policies. LERC~\cite{8254999} extended this with coordinated eviction respecting the all-or-nothing property, saying that a task only benefits from caching if all input dependencies remain in memory. Similar approaches optimize materialized views and distributed query execution~\cite{10184758,sioulas2023realtimeanalyticscoordinatingreuse}. These systems target batch workloads with static, data-centric DAGs.
\paragraph{Multi-Agent Systems with DAG Planning.}
Agent systems adopt DAGs to structure task execution, where nodes represent agent actions (LLM calls, tool invocations, computations) and edges encode execution dependencies. Recent work demonstrates how agents coordinate through DAG based task graphs for subject specific reasoning and decentralized architectures~\cite{dong2025sdagsubjectbaseddirectedacyclic,yang2025agentnet}, while benchmarks characterize common DAG patterns in multi agent workflows~\cite{qiao2025benchmarking}. 
\subsection{Multi-Agent System Orchestration and Optimization}
Beyond just caching, there is also research that explores complementary approaches to improving multi-agent efficiency through orchestration, scheduling, and workflow design.
\paragraph{Scheduling and Orchestration.}
There are several systems that optimize agent execution through workflow aware resource management. Kairos~\cite{chen2025kairoslowlatencymultiagentserving} introduces priority scheduling that accounts for agent execution characteristics and memory constraints. Ayo~\cite{tan2025ayo} is another example that proposes fine grained dataflow orchestration, which can enable cross module optimizations. These approaches improve efficiency by optimizing execution order and resource allocation and often operate orthogonally to caching.
\paragraph{Workflow Generation and Automation.}
Recent work has also explored automated workflow construction. AFlow~\cite{zhang2025aflow} and ADAS~\cite{hu2025automated} learn effective collaboration patterns and agents by searching over and learning about workflow architectures. DynaSaur~\cite{nguyen2024dynasaur} enables dynamic action composition, allowing agents to adaptively modify workflows based on intermediate results.
These orchestration and optimization techniques address concerns orthogonal to caching. Scheduling focuses on execution ordering, workflow generation improves task decomposition, while caching eliminates redundant computation. The techniques can be combined. Better scheduling may improve cache locality, while effective caching reduces computational load.
\section{Design}
\label{sec:method}
Multi-agent workflows decompose complex tasks into sequences of specialized operations, such as retrieval, reasoning, synthesis, and more. These operations can form directed acyclic graphs (DAGs) where each node represents an agent execution and edges encode dependencies between tasks. Caching expensive operations from this DAG such as LLM reasoning, visual encoding, and OCR can dramatically reduce latency by avoiding redundant computation across queries. However, existing eviction policies miss three signals unique to multi-agent workloads; DAG topology, which reveals which cached results feed downstream tasks, execution costs which can vary by orders of magnitude across agent types, and agent invocation frequency which can also reflect how likely a cached result is to be reused. We present a workload-aware eviction policy that combines these three signals into a unified scoring function to maximize cache utility for multi agent systems under memory constraints.

\subsection{Problem Formulation}
Consider a workflow executor processing queries with a multi agent system by decomposing each into a task DAG. Each task executes using a specialized agent and produces a result consumed by dependent tasks downstream.

\paragraph{Task DAG} For query $q$, the workflow is a DAG, $G = (V, E)$ where $V = \{t_1, \ldots, t_n\}$ are tasks and $(t_i, t_j) \in E$ means $t_j$ depends on $t_i$'s result. Each task has an associated agent type $a(t_i)$, such as retrieval, OCR, or synthesis.

\paragraph{Cache Model} Task execution results are stored as keyed mappings under a fixed capacity $C$. Cache entries are keyed by a tuple of the task description and subject identifier (e.g., the source document or video) explicitly outputted by the planning agent. When capacity is exceeded, the eviction policy selects a victim entry to remove.

\paragraph{Execution Costs.} Each task $t_i$ incurs recomputation time $\tau(t_i)$ on a cache miss, varying by agent type and operation. A cache hit saves $\tau(t_i)$, and a miss pays it.

\subsection{Workload Aware Cache Management}

LRU and LFU rely exclusively on access history (recency or frequency), blind to future reuse and cost heterogeneity. Our approach combines complementary signals.

\subsubsection{Dependency Count} 
DAG topology allows for forward-looking reuse estimation. For a cached task $t_i$, we track $D(t_i)$, the set of downstream tasks that consume its result 
\[ D(t_i) = \{t_j : (t_i, t_j) \in E\} \] 

The dependent count $|D(t_i)|$ serves as a structural proxy for generality and future value. An entry with zero dependents feeds nothing downstream and is generally safe to evict. Entries with many dependents, by contrast, are consumed by a larger number of distinct child tasks, which shows that their result may be general enough to be useful across many different similar computational contexts. Nodes that accumulate substantial upstream context through prior steps (previous steps acting as something of an informational filter) tend to produce results that are less generalizable, as their output is increasingly specific to the particular query path taken. Frame extraction in a video analysis workload is a natural example, where a single extraction result feeds different downstream analysis agents independently. By contrast, a final answering node at the bottom of the same DAG has zero dependents and is maximally query specific. While this relationship is not absolute, dependent count provides an intuitive and lightweight structural signal for how broadly reusable a cached entry is likely to be, complementing the recomputation cost and frequency signals in the combined scoring function.

\subsubsection{Recomputation Cost} 
Not all cache misses carry equal penalties. The cost of a miss is determined entirely by how long the evicted entry takes to recompute, and in heterogeneous agentic workloads this variation can be substantial. Evicting an agentic operation taking on the order of 8 to 12 seconds, is far more damaging than evicting a lightweight answering step that completes in under a second. Yet traditional policies like LRU treat these entries identically, making eviction decisions purely on access patterns regardless of the computational investment each entry represents. We record the observed execution time $\tau(t_i)$ for each cached entry and use it directly as a scoring signal, so that the policy is explicitly aware of what it would cost to recompute each entry on a miss. This directly addresses a shortcoming of cost-blind policies where an entry accessed slightly less recently but requiring much more computation to recompute should not be evicted simply because another entry was touched more recently.

\subsubsection{Agent Invocation Frequency} 
Beyond the structure of individual DAGs, the distribution of agent invocations across the query stream carries its own signal about future reuse. An agent type that has been invoked frequently up to the current point in the workload is likely to remain active. If agent A has been called 120 times while an agent B has been called only 20 times, the workload is clearly oriented toward agent A's content, and entries produced by frame extraction are correspondingly more likely to be needed again. We track the cumulative invocation count $f(a)$ for each agent type $a$ across all queries processed so far and use it as a proxy for expected future reuse. Entries from high-frequency agents receive higher keep scores, while entries from rarely invoked agents are deprioritized. This allows the policy to naturally be more inclined towards keeping agent types that are central to the current workload focus, such as a database query agent in a financial analysis workload where many queries retrieve information from the same set of reports, without requiring any explicit workload specification from the user.

\subsubsection{Combined Score}
The three signals combine into a single keep score:
\[
\text{score}(t_i) = w_{\text{dep}} \cdot D(t_i) + w_{\text{cost}} \cdot \tau(t_i) + w_{\text{freq}} \cdot f(a(t_i))
\]
where $w_{\text{dep}}$, $w_{\text{cost}}$, and $w_{\text{freq}}$ are non-negative weights. The eviction policy removes the entry minimizing this score, which is the entry with fewest dependents, lowest recomputation cost, and least-invoked agent type. Section~\ref{sec:evaluation} examines weight sensitivity with ablation experiments.

\begin{figure}[t]
\centering
\includegraphics[width=0.75\columnwidth]{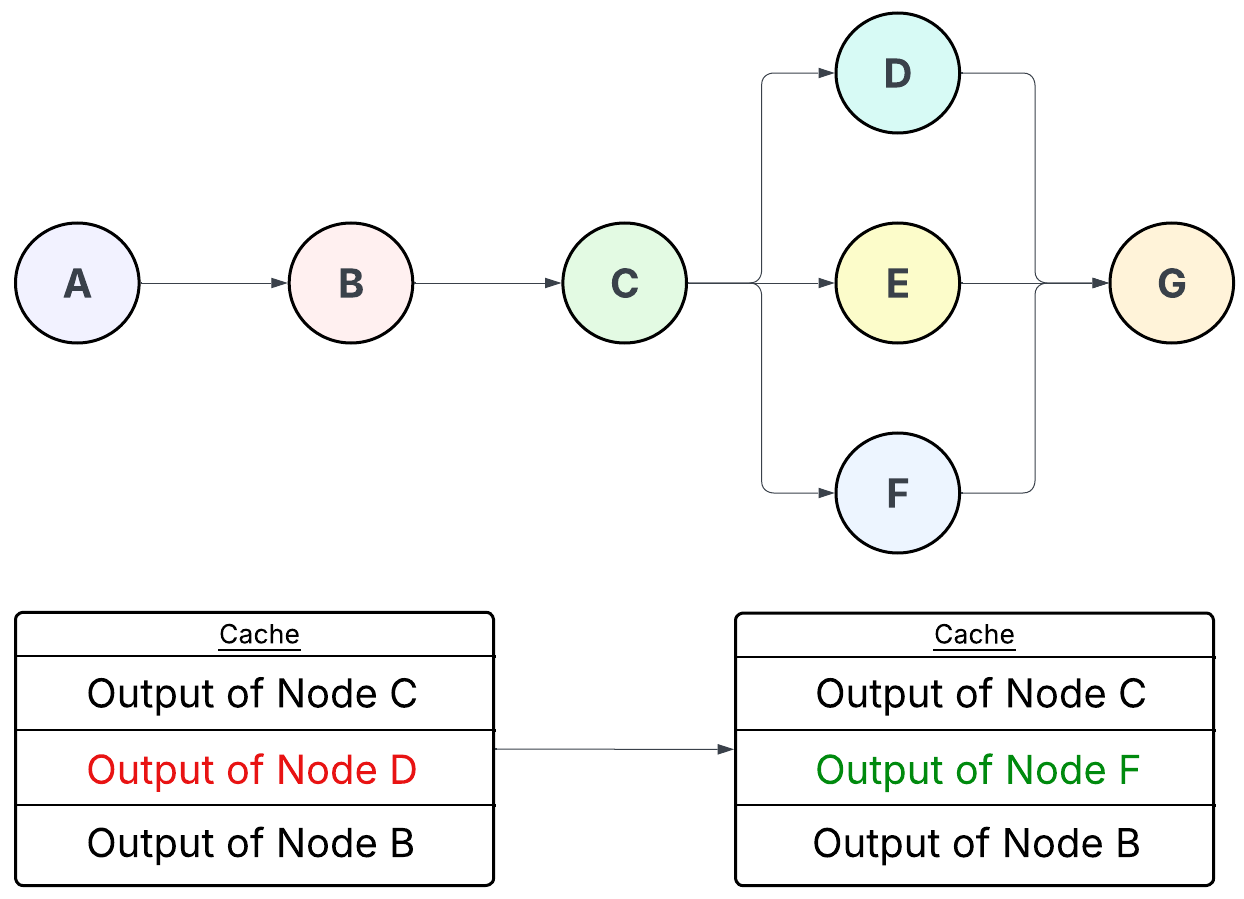}
\caption{Workload aware eviction decision. When cache capacity is exceeded, our 
policy takes into account workload characteristics such as dependency count, execution cost, and agent frequency when making decisions. Node D is correctly identified 
as lower priority despite being recently accessed, avoiding premature 
eviction of higher value nodes such as Node C.}
\label{fig:eviction-example}
\end{figure}

\subsection{Eviction Policy}

We maintain cached entries in a min-heap ordered by keep score. Algorithm~\ref{alg:eviction} describes the procedure. When a task result is cached, we compute its score based on current workload state with the signals of dependency count, execution cost, and agent frequency, and insert it into the heap. When capacity is exceeded, eviction is done in $O(\log n)$ time. 

\begin{algorithm}
\caption{Workload-Aware Cache Eviction}
\label{alg:eviction}
\begin{algorithmic}[1]
\Require Cache capacity $C$, min-heap $H$, weights $w_{\text{dep}}, w_{\text{cost}}, w_{\text{freq}}$
\State
\Function{CacheInsert}{task $t_i$, result, DAG $G$}
    \State $\text{score} \gets w_{\text{dep}} \cdot D(t_i) + w_{\text{cost}} \cdot \tau(t_i) + w_{\text{freq}} \cdot f(a(t_i))$
    \State $H.\text{insert}(t_i, \text{score})$ \Comment{$O(\log n)$}
    \While{cache size $> C$}
        \State $\text{victim} \gets H.\text{extract-min}()$ \Comment{$O(\log n)$}
        \State Evict victim
    \EndWhile
\EndFunction
\end{algorithmic}
\end{algorithm}
\section{Evaluation} 
\label{sec:evaluation}
We evaluate our workload aware eviction policy against standard cache replacement baselines across a diverse set of multi agent workloads. Our key findings are: 

\begin{itemize} 

\item \textbf{Comparison to Unbounded Caching} Our workload aware policy consistently approaches the performance of the unbounded cache across benchmarks, achieving latency within a small margin of the theoretical upper bound attainable under unbounded memory (\S\ref{sec:cache_management}). 

\item \textbf{Superior efficiency over all other finite-capacity baselines.} Workload aware caching achieves on average a 31.1\% reduction in latency over the next best finite capacity baseline, and reduces latency by up to 64.7\% relative to the uncached baseline across benchmarks (\S\ref{sec:cache_management}). 

\item \textbf{Substantial throughput gains.} Our policy significantly increases system throughput across all benchmarks, achieving up to a 2.84$\times$ improvement over the uncached baseline (\S\ref{sec:cache_management}). 

\item \textbf{Accuracy.} Workload aware caching retains on average 94.7\% of the uncached baseline's accuracy across benchmarks, while matching or surpassing the accuracy of all competing finite-capacity methods. (\S\ref{sec:cache_management}). 

\end{itemize}

\subsection{Experimental Setup}
\subsubsection{Benchmarks}
We evaluate on three benchmarks including document, presentation, and video understanding tasks.
\paragraph{SlideVQA} A presentation question answering benchmark~\cite{SlideVQA2023} where queries target slides within presentation decks. Workloads include 1-13 queries per deck. This represents \textbf{moderate reuse}, where we have multiple queries with access to the same deck. This means creating opportunities for caching different results from the DAGs. However, the final answers remain query specific, even if evidence from previous steps are similar.

\paragraph{MP-DocVQA} A multi page document QA benchmark~\cite{tito2023hierarchicalmultimodaltransformersmultipage} where queries are about information in PDF documents with evidence found throughout pages. Each document contains 1-20 pages, with 1-85 structurally similar queries per document. This represents \textbf{highly variable reuse}, where queries have extensive reuse of operations as users explore documents in depth across many different queries in some cases. There are many documents for which there are little to no reusable operations, and also many for which there is an extensive amount of reuse.

\paragraph{VideoMME} A video understanding benchmark~\cite{fu2025video} where queries are about content in video files. Each video receives 3 queries  about content (e.g temporal, visual). This represents \textbf{low reuse}, where limited reuse on similar videos results in higher cache churn. Sub-tasks are highly, highly variable in terms of agent costs.

\subsubsection{Multi-Agent Workflows}
All benchmarks execute using a multi agent system, implemented as a Plan-Act architecture\cite{erdogan2025planandact}. Given a query, a planning agent (using Qwen2.5:14b as a model backbone) decomposes it into a DAG of tasks needed to answer it, each assigned to a specialized agent. For example, a frame extraction agent, an OCR agent, a visual analysis agent, etc. These agents are quite heterogeneous in terms of their execution times and complexity. Specialized agents utilize the Qwen2.5:7b model. The final answers to queries are synthesized and reasoned about with the Qwen2.5:14b model. 

\subsubsection{Baseline Eviction Policies}
We compare against standard cache replacement policies, including:
\begin{itemize}
\item \textbf{No Cache:} Execute all tasks without caching (upper bound on execution time)
\item \textbf{LRU (Least Recently Used):} Evicts the entry accessed least recently
\item \textbf{LFU (Least Frequently Used):} Evicts the entry accessed least frequently  
\item \textbf{FIFO (First-In-First-Out):} Evicts the oldest entry
\item \textbf{ARC (Adaptive Replacement Cache~\cite{270366}):} Dynamically balances recency and frequency using two lists
\item \textbf{Workload Aware (Ours):} Utilizes workload characteristics for eviction decisions as described in Section~\ref{sec:method}
\end{itemize}

\paragraph{Agentic Optimization Baselines} 
Beyond standard eviction policies, we also compare against complementary agentic optimization techniques in Section~\ref{sec:agentic_opt}, including caching via GPTCache~\cite{bang-2023-gptcache}, plan-level caching via Agentic Plan Caching~\cite{zhang2025agentic}, and parallel agent execution~\cite{kim2024llmcompiler}. These methods target different bottlenecks in the agentic pipeline.

\subsubsection{Hardware and Experimental Parameters}
\begin{figure}[htbp]
    \centering
    \includegraphics[width=0.9\linewidth]{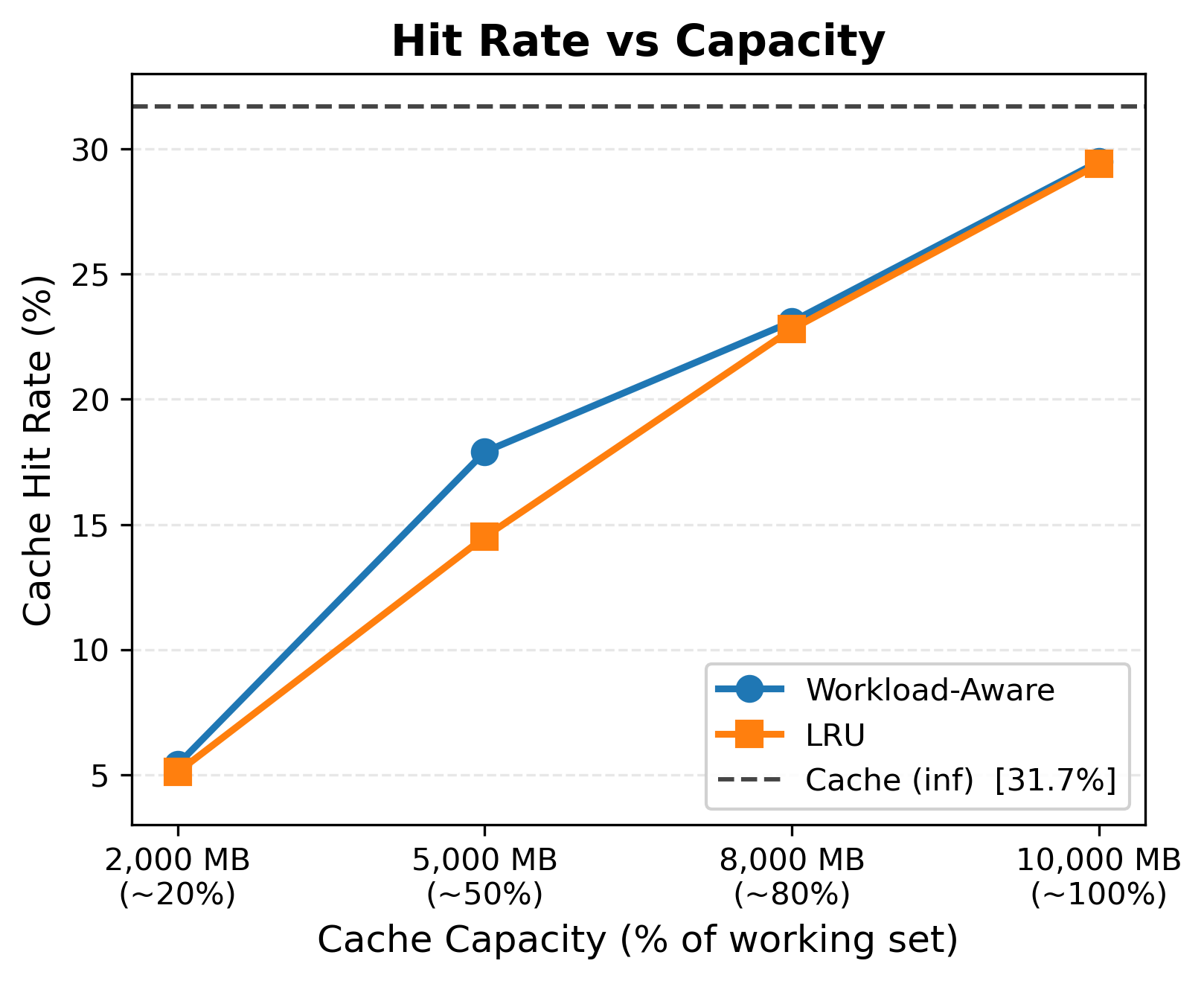}
    \vskip\baselineskip
    \includegraphics[width=0.9\linewidth]{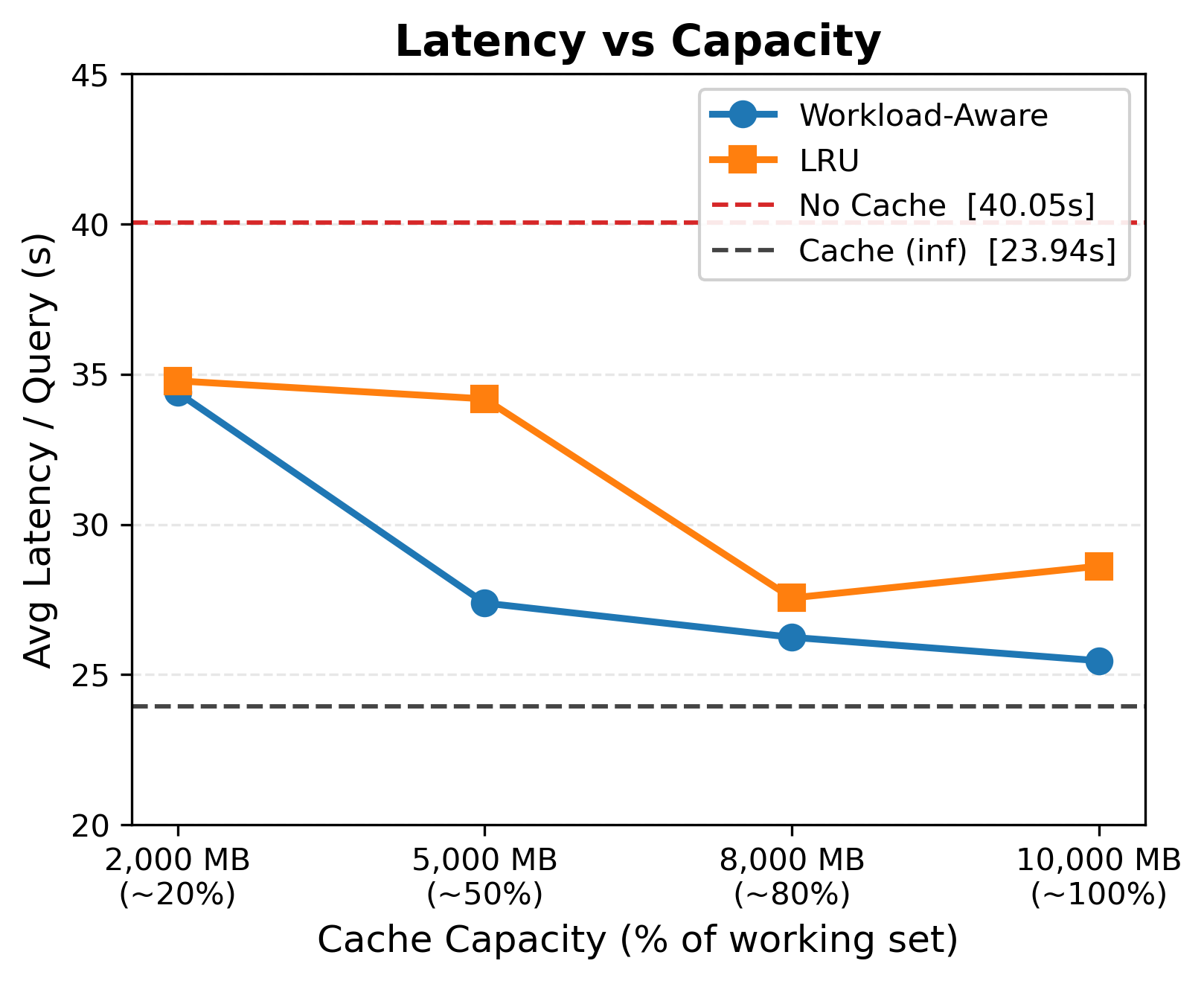}
    \caption{Cache hit rate (top) and latency (bottom) across capacity levels on VideoMME, comparing Workload-Aware and LRU eviction policies. The largest performance gap is seen at a midpoint capacity (~50\% of working set), where eviction decisions have the most impact. Policies converge as capacity approaches the full working set.}
    \label{fig:capacity}
\end{figure}

We ran experiments in an HPC environment with a single A100 GPU, using the Qwen2.5:14b and Qwen2.5:7b models~\cite{qwen2025} for our agents through use of Ollama~\cite{Ollama}. Scoring weights were set to $w_{\text{dep}} = 1$, $w_{\text{cost}} = 2$, and $w_{\text{freq}} = 0.01$, held constant across all benchmarks for fairness. Execution cost is weighted most heavily to account for the high heterogeneity in agent execution times, while the smaller frequency weight allows cumulative invocation counts to increase the signal's contribution as agents are used more across queries. The optimal weights may differ depending on observed workload characteristics, and adaptive weight selection is an interesting direction for future work. Cache capacity is set to 50\% of a working set round for each benchmark, where a round represents the minimum query volume before the cache begins seeing repeated source/subject material. This maps to 5,000MB for VideoMME and 500MB for SlideVQA and MP-DocVQA. To validate this choice, we sweep across capacity levels on VideoMME, comparing our policy against LRU. As shown in Figure~\ref{fig:capacity}, the largest differences between policies occurs at moderate capacity, where eviction decisions have meaningful impact. At lower capacities, the eviction pressure is too high for policies to get useful hits. At higher capacities, policies converge as eviction becomes rare/the eviction pressure is much too low. The 50\% capacity choice is a good midpoint which allows for a reasonable amount of eviction pressure where policy choices actually matter, and we can see the differences between them more clearly.

\subsection{Cache Management} \label{sec:cache_management} 
\subsubsection{Main Results} 
\begin{figure*}[htbp]
    \centering
    \includegraphics[width=\textwidth]{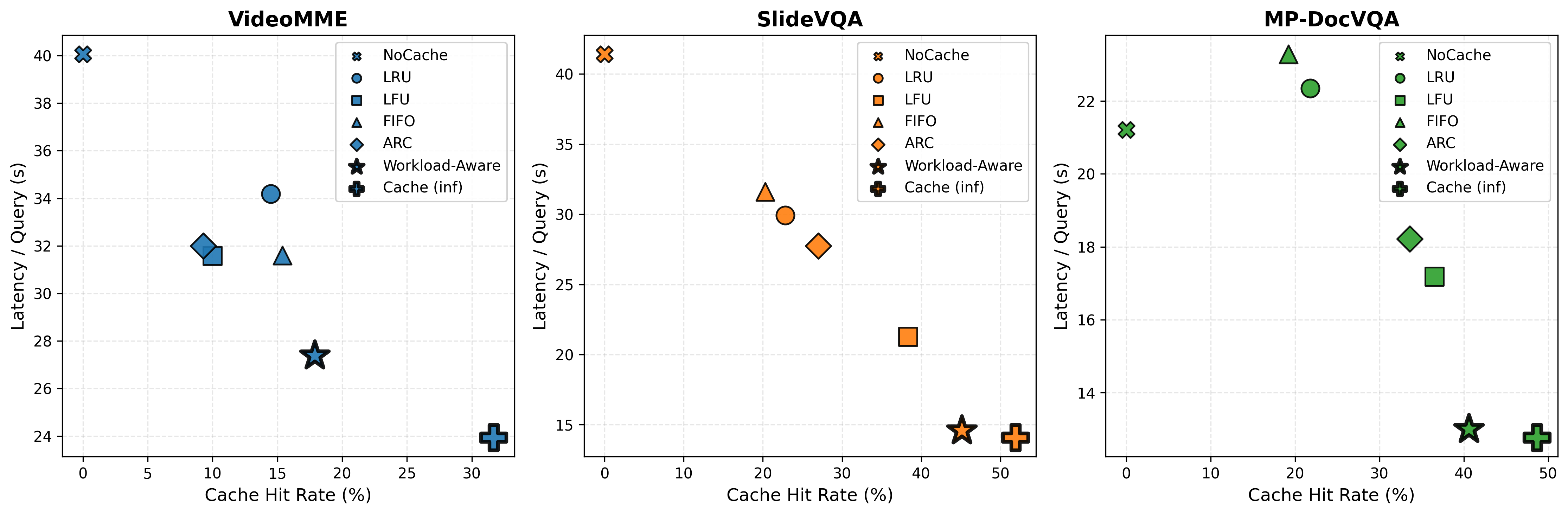}
    \caption{Latency vs Hit Rate. Demonstrates that Workload Aware caching consistently outperforms other baselines in both metrics, and often approaches the performance of an unbounded cache.}
    \label{fig:scatter_wide}
\end{figure*}

\begin{table*}[t]
\centering
\caption{Performance comparison across datasets. Note that MP-DocVQA and SlideVQA evaluate plain English answers using the Average Normalized Levenshtein Similarity (ANLS) metric, whereas VideoMME uses accuracy percentages for multiple-choice questions.}
\label{tab:combined_results_wide}
\begin{tabular}{l ccc ccc ccc}
\toprule
\textbf{Method} & \multicolumn{3}{c}{\textbf{MP-DocVQA}} & \multicolumn{3}{c}{\textbf{SlideVQA}} & \multicolumn{3}{c}{\textbf{VideoMME}} \\
\midrule
& \textbf{Hit Rate} & \textbf{Lat (s)} & \textbf{Acc (ANLS)} & \textbf{Hit Rate} & \textbf{Lat (s)} & \textbf{Acc (ANLS)} & \textbf{Hit Rate} & \textbf{Lat (s)} & \textbf{Acc (\%)} \\
\midrule
Uncached Baseline     & ---    & 29.99 & 0.547 & ---    & 41.40 & 0.500 & ---    & 40.05 & 35.4\% \\
Infinite Cache        & 48.7\% & 12.77 & 0.547 & 51.9\% & 14.09 & 0.436 & 31.7\% & 23.94 & 31.5\% \\
\midrule
Workload-Aware (Ours) & \textbf{40.6\%} & \textbf{13.00} & \textbf{0.548} & \textbf{45.1\%} & \textbf{14.59} & 0.451 & \textbf{17.9\%} & \textbf{27.38} & 33.2\% \\
LFU                   & 36.5\% & 17.19 & 0.547 & 38.3\% & 21.27 & 0.439 & 10.0\% & 31.58 & 30.9\% \\
ARC                   & 33.6\% & 18.22 & 0.546 & 27.0\% & 27.75 & 0.461 & 9.3\%  & 31.99 & 31.3\% \\
LRU                   & 21.8\% & 22.35 & 0.545 & 22.8\% & 29.93 & 0.466 & 14.5\% & 34.18 & 30.1\% \\
FIFO                  & 19.2\% & 23.28 & 0.546 & 20.3\% & 31.61 & 0.480 & 15.4\% & 31.59 & 31.1\% \\
\bottomrule
\end{tabular}
\end{table*}
Table~\ref{tab:combined_results_wide} presents our main results across 3 benchmarks. Our workload aware policy consistently achieves the highest hit rates and lowest latency while maintaining accuracy in the realm of the other caching eviction baselines. Under a fixed memory budget, it often approaches the performance of an infinite memory/unbounded cache. Other methods often come with both a lower hit rate and lower accuracy, with the exception of the SlideVQA dataset, where the differences in hit rate and latency are most stark in favor of workload aware caching. The scatter plots in Figure~\ref{fig:scatter_wide} further demonstrate the benefits of Workload Aware caching.

On workloads with moderate to high reuse opportunity, we see the biggest differences in terms of hit rates and latency, as the cache has more flexibility to achieve hits. On SlideVQA and MP-DocVQA, we achieve hit rates of 45.1\% and 40.6\% respectively, reducing per query latency by 64.7\% and 56.6\% relative to the uncached baseline, and improving throughput by up to 2.84$\times$. It is also important to note that these gains are at a minimal distance from the theoretical upper bound of a cache with infinite storage capacity. Our policy achieves latency within 3.4\% and 1.8\% of an infinite cache on SlideVQA and MP-DocVQA respectively, despite having a fixed memory budget. The next best finite capacity approach, LFU, is behind our approach by 45.8\% and 32.2\% on the same benchmarks. ARC, LFU, and FIFO perform progressively worse. For instance, LRU and FIFO on MP-DocVQA achieve hit rates of just 21.8\% and 19.2\% respectively, with latency even approaching that of the uncached baseline.

On datasets with lower reuse, such as VideoMME, Workload Aware caching continues to outperform other finite capacity baselines. With only approximately three queries per video, the cache has limited opportunity to accumulate reusable results before transitioning to a different video with a distinct profile for reusing agent steps. This can be seen with the universally lower hit rates across all policies. Our workload aware policy achieves the highest hit rate at 17.9\%, as well as the lowest latency of 27.38 seconds per query amongst finite capacity methods. This represents a 31.6\% latency reduction when compared to the uncached baseline.

In summary, across all benchmarks, our workload aware policy maintains or surpasses the accuracy of all competing finite capacity methods, and in several cases closely approaches the uncached baseline among all caching methods.

\subsubsection{Eviction Cost Analysis}
\begin{figure}[htbp]
    \centering
    \includegraphics[width=\linewidth]{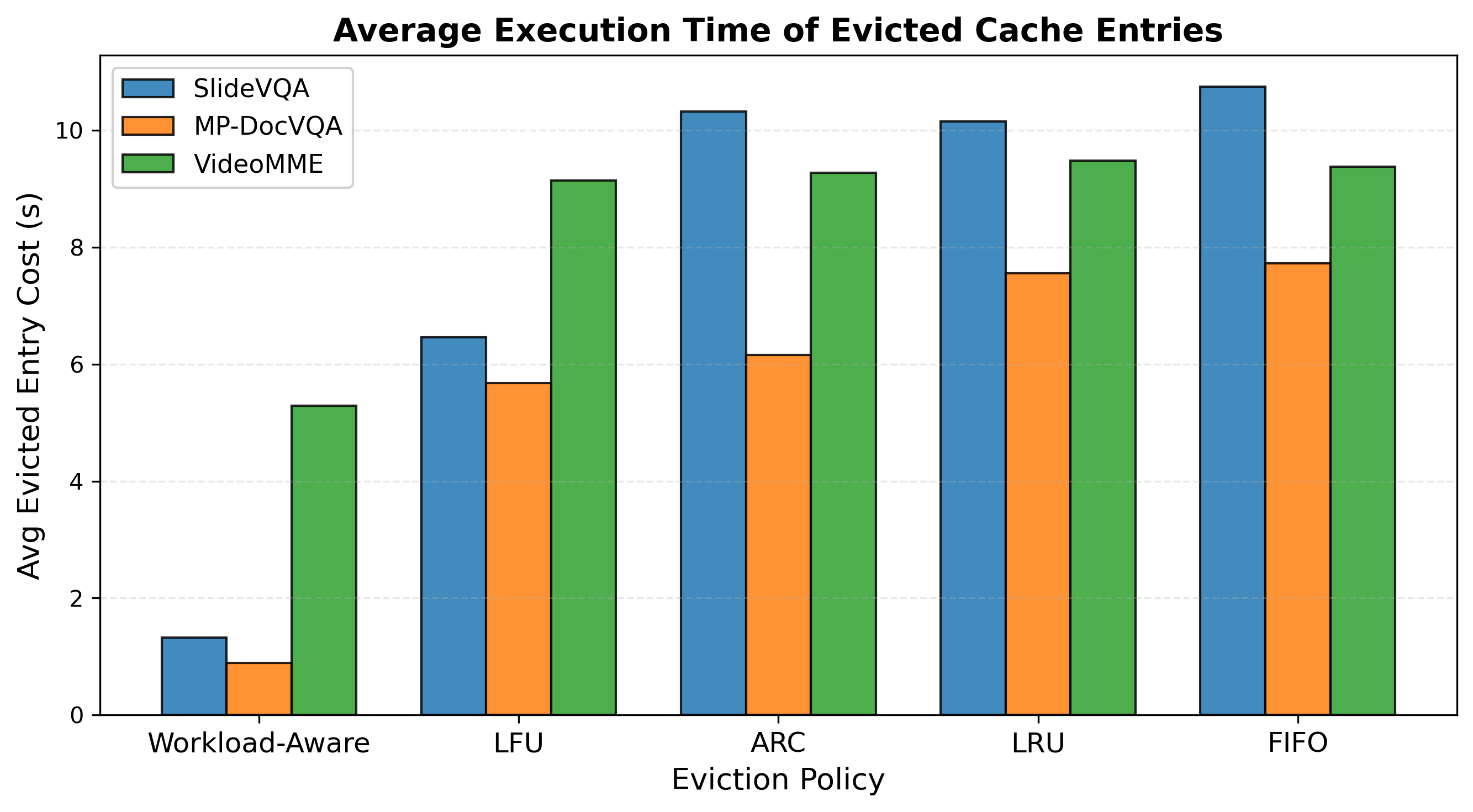}
    \caption{Average Evicted Cost of Entries for Each Method}
    \label{fig:evicted_cost}
\end{figure}
\begin{table}[htbp]
\centering
\caption{Execution time saved by cache hits as a percentage of the time saved by Workload-Aware caching, which saves the most cost out of all methods. Higher is better.}
\label{tab:cost_saved}
\begin{tabular}{lccc}
\toprule
Method & SlideVQA & MP-DocVQA & VideoMME \\
\midrule
LFU & 74.3\% & 75.7\% & 42.2\% \\
ARC & 54.0\% & 70.1\% & 40.9\% \\
LRU & 46.9\% & 44.7\% & 62.4\% \\
FIFO & 42.2\% & 39.4\% & 68.0\% \\
\bottomrule
\end{tabular}
\end{table}

Cache hit rate, while an important metric, does not always tell the full story of performance~\cite{qiu2024increasinghitratiohurt}. Figure~\ref{fig:evicted_cost} and Table~\ref{tab:cost_saved} demonstrate one reason for our method's increase in serving performance. Workload aware caching has the capability to identify costly entries in the cache and preserve them, which means the evicted items are dramatically less costly than what the other policies evict on all three benchmarks. Cost blind policies such as LRU, FIFO, LFU, and ARC evict purely based on access history, making them just as likely to evict an expensive frame extraction or OCR result as a cheap lightweight lookup operation. With recomputation cost added as a scoring signal as in our policy, we are able to ensure expensive items remain in the cache.

The practical consequences of this are also very significant. When a cache miss does occur with our policy, it tends to be for a cheaper entry, while misses on other policies have a higher probability of being for an expensive node/agentic step. This asymmetry compounds over time, and as shown in Table~\ref{tab:combined_results_wide}, our policy ends up with a lower query latency amongst all benchmarks compared to the uncached baseline and all finite capacity methods, despite the absolute difference in hit rate not always being large. This indicates that the quality of what is kept in the cache matters just as much as how often the cache is hit, especially with heterogeneous workflows such as agentic ones.  

\subsubsection{Cross-Model Results}
\begin{figure*}[htbp] 
    \centering 
    \includegraphics[width=\textwidth]{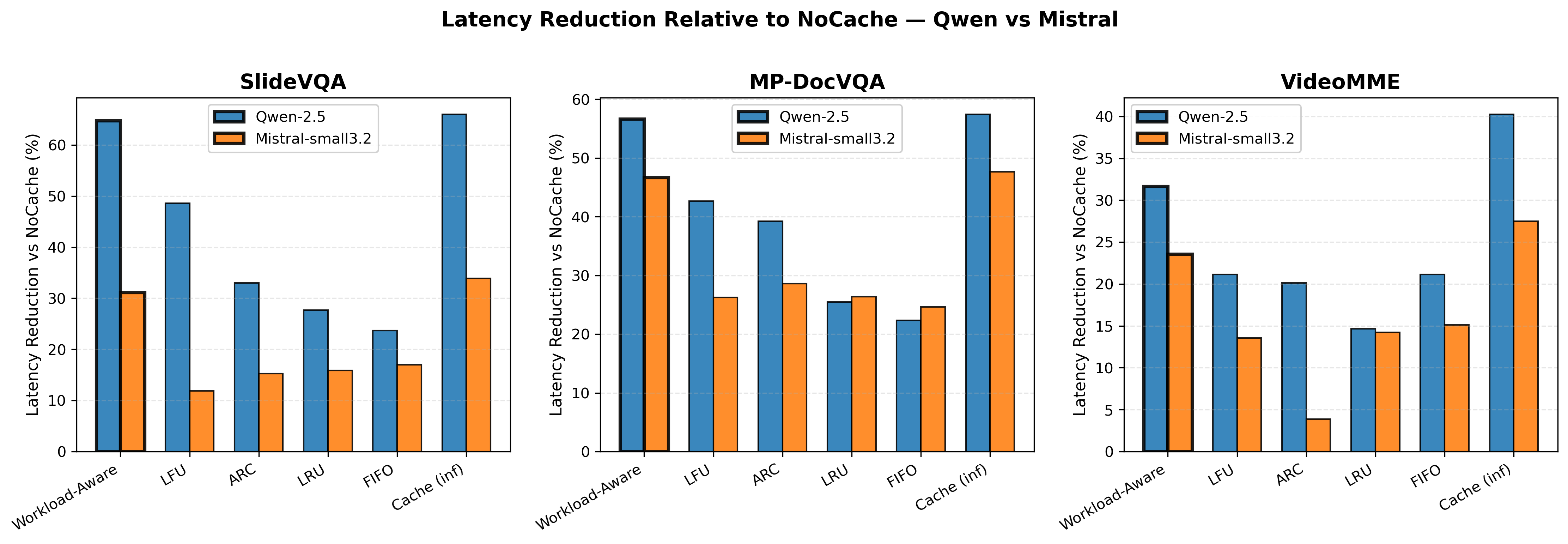} \caption{Latency reduction relative to the uncached baseline across all methods and benchmarks, comparing Qwen-2.5 and Mistral-small3.2 model families. Workload-Aware caching consistently achieves the highest latency reduction among finite-capacity methods across both model families, demonstrating that the policy's advantage is not specific to a particular model. Absolute latency reductions differ between model families due to differences in per-query inference speed and planning behavior, but the relative ordering of policies is preserved.} \label{fig:generalization}
\end{figure*}
We additionally evaluate our policy's performance by replicating the main experiments using Mistral-small3.2:24b\cite{mistralsmall32} as the agent backbone across all three benchmarks. Figure~\ref{fig:generalization} shows latency reduction relative to the uncached baseline for both model families. Workload-Aware caching consistently achieves the highest latency reduction among finite-capacity methods across all three benchmarks and both model families, with the relative ordering of policies preserved in all cases. Absolute latency reductions differ between model families due to differences in per-query inference speed and planning behavior, but the advantage of Workload-Aware caching over competing baselines is robust across both settings.

\subsection{Agentic Optimization} 
\label{sec:agentic_opt} 
\begin{table}[htbp]
\centering
\caption{Performance comparison of agentic baselines, highlighting latency and accuracy trade-offs. Accuracy is reported in ANLS for SlideVQA and standard percentage for VideoMME.}
\label{tab:agentic_baselines}
\begin{tabular}{lcc}
\toprule
\textbf{Method} & \textbf{Latency (s)} & \textbf{Accuracy} \\
\midrule
\multicolumn{3}{c}{\textbf{SlideVQA (ANLS)}} \\
\midrule
Workload-Aware (ref)    & 14.59 & 0.451 \\
Parallel    & 40.08 & 0.500 \\
APC (cold)      & 42.30 & 0.495 \\
APC (warm)      & 38.69 & 0.479 \\
GPTCache        & 13.87 & 0.391 \\
APC+Parallel+Workload-Aware (cold) & 30.12 & 0.398 \\
APC+Parallel+Workload-Aware (warm) & 21.54 & 0.425 \\
\midrule
\multicolumn{3}{c}{\textbf{VideoMME (\%)}} \\
\midrule
Workload-Aware (ref)    & 27.38 & 33.2\% \\
Parallel        & 30.96 & 34.9\% \\
APC (cold)      & 34.49 & 36.0\% \\
APC (warm)      & 32.86 & 36.8\% \\
GPTCache        & 25.02 & 30.4\%  \\
APC+Parallel+Workload-Aware (cold) & 27.71 & 35.2\% \\
APC+Parallel+Workload-Aware (warm) & 27.42 & 34.6\% \\
\bottomrule
\end{tabular}
\end{table}
We evaluate how our workload-aware caching interacts with two complementary multi-agent efficiency techniques, each targeting different bottlenecks in multi-agent systems. First, parallel execution of agents similar to LLMCompiler~\cite{kim2024llmcompiler}, executing independent DAG nodes concurrently instead of sequentially to reduce latency. For plan-level caching, we compare with Agentic Plan Caching (APC) ~\cite{zhang2025agentic}, which reuses structured plan templates across tasks to reduce planning overhead. Our technique, in comparison, targets the bottleneck of redundant node level recomputation across queries. We evaluate each technique in isolation and in combination with our approach. For APC, we report both cold start results, where the cache starts empty, and warm results. Results are shown in Table~\ref{tab:agentic_baselines}. Results are reported for benchmarks with moderate reuse (SlideVQA) and low reuse (VideoMME) to demonstrate the performance of each method in different settings.

Our workload aware caching achieves the best latency as a stand alone method, outperforming other individual techniques. Alone, parallel execution provides notable latency benefits depending on the workload. For example, on VideoMME, where the opportunities for overlap are high (e.g. captioning frames and extracting audio are independent of each other, yet both computationally expensive) parallelism has an obvious impact relative to the uncached baseline (see Table~\ref{tab:combined_results_wide}). In comparison, on SlideVQA, the benefits of parallelism can be seen but are smaller relative to the uncached baseline, as the dependencies between steps differ, and the latency attributed to them may be more heterogeneous (thereby reducing the impact of parallelism). 

APC alone shows limited impact under cold start conditions, where the overhead of cache generation offsets much of the potential planning savings in our workloads. With a lower number of repeated/heavily related queries, the benefits of caching plans are not seen immediately, resulting in cold start latency that is equal to or slightly worse than the uncached baseline. As the template cache warms up, APC begins to demonstrate more meaningful improvement. Warm APC results show reduced latency relative to cold on both benchmarks, confirming plan level reuse is a useful signal to use alongside others, especially with a populated cache. 

GPTCache style semantic caching shows a double edged tradeoff, leading to lower latency on both benchmarks, but at a consistent accuracy cost across both benchmarks as well. This occurs because approximate matches return cached intermediate results that are incorrect for the current query, propagating errors through downstream agent steps and correspondingly lowering the accuracy. This reflects a characteristic of agentic settings, as unlike end-to-end query caching where a semantically similar response may be acceptable, intermediate DAG node results must be precise since they feed into dependent computations. Exact match caching avoids this tradeoff, prioritizing correctness over the raw hit rate.

The combination of APC, parallel execution, and workload-aware content caching was also evaluated, and shows clear synergy. Each technique targets a distinct stage of the agentic pipeline, and together they can collectively reduce latency across planning, execution, and node-level recomputation. On VideoMME, APC + Parallel + Workload-Aware achieves competitive latency with our standalone approach in cold conditions, and warm results show further improvement as plan templates accumulate. These results suggest that the three techniques are genuinely complementary, with each contributing independently to overall efficiency. On SlideVQA, however, the combined approach shows higher latency than our standalone workload-aware caching. This reflects a practical deployment consideration in resource-constrained environments. In our case, running parallel agent execution (so potentially multiple GPU-bound tools concurrently) alongside language model inference and cached items (e.g. tensors) on a single GPU introduces resource contention that can degrade throughput.

Overall, these results highlight that workload-aware content caching, plan-level caching, and parallel execution are complementary techniques that each address a distinct efficiency bottleneck in multi-agent systems. Our approach performs strongly as a standalone method and provides a solid foundation that other techniques can build upon.

\subsection{Ablation Study}
\label{sec:ablation}
\begin{table}[htbp]
\centering
\caption{Ablation study of the Workload-Aware scoring function components. Accuracy is reported in ANLS for SlideVQA and standard percentage for VideoMME.}
\label{tab:ablations}
\begin{tabular}{lccc}
\toprule
\textbf{Method} & \textbf{Hit Rate} & \textbf{Latency (s)} & \textbf{Accuracy} \\
\midrule
\multicolumn{4}{c}{\textbf{SlideVQA (ANLS)}} \\
\midrule
Workload-Aware (full) & 45.1\% & 14.59 & 0.451 \\
No-Freq    & 40.1\% & 23.39 & 0.427 \\
No-Dep     & 38.0\% & 23.50 & 0.443 \\
No-Exec    & 1.4\%  & 41.88 & 0.498 \\
\midrule
\multicolumn{4}{c}{\textbf{VideoMME (\%)}} \\
\midrule
Workload-Aware (full) & 17.9\% & 27.38 & 33.2\% \\
No-Freq    & 15.6\% & 32.15 & 31.7\% \\
No-Exec    & 15.3\% & 33.04 & 29.8\% \\
No-Dep     & 14.7\% & 31.47 & 30.6\% \\
\bottomrule
\end{tabular}
\end{table}
To evaluate the choice of parameters in our workload aware policy, we ablate the three scoring signals. We evaluate three variants, leaving one of the scoring signals out each time. Results are shown in Table~\ref{tab:ablations} for SlideVQA and VideoMME. 

Across both benchmarks, removing any single scoring signal degrades hit rate and latency relative to utilizing all three, confirming that the combination of signals is robustly beneficial across different workloads. 

The relative importance of each signal differs across workloads. On SlideVQA, removing the execution time term has by far the largest impact, collapsing hit rate and degrading latency to near the uncached baseline. This dramatic change reflects the important role of execution cost as a protective signal in this workload, indicating that items that were expensive were also more likely to be reused there. Removing frequency or dependency count produces a more moderate, yet still noticeable degradation in comparison to the full policy. 

On VideoMME, all three ablations degrade performance by comparable margins, with hit rates falling from 17.9\% to between 14.7 and 15.6\%, and average query latency increasing from 27.38s to around 31-34 seconds. Here, dependency count carried the largest independent impact, with No-Dep having the lowest hit rate. The execution time term, while still important, does not dominate to the same degree as it did in SlideVQA, indicating a difference in the costs of agents, which ones are reused most, and the way they are used in the planning DAGs. 

Taken all together, the ablation results validate our choices of parameters. Each signal addresses a distinct aspect of eviction quality. Execution cost protects expensive entries, while dependency count and agent frequency reflect workload alignment (e.g. agents used quite frequently or entries that have many dependencies are more likely to be reused in the workload). The full combination consistently outperforms other subsets, and the workload dependent variation in signal importance motivates the use of all three rather than a single one.
\section{Limitations and Future Work}

Our current evaluation uses fixed weights for the three scoring signals, $w_{\text{dep}}$, $w_{\text{cost}}$, and $w_{\text{freq}}$, which are held constant across the query stream. While our ablation study shows that all three signals contribute meaningfully across diverse workloads, the relative importance of each signal varies by workload characteristics. For instance, a workload with fairly consistent agent-calling frequencies but heavily variable execution times between steps might work better with a lower frequency signal. An interesting direction for future work is to learn or adapt these weights dynamically based on similar observed query patterns, allowing the policy to specialize to the workload distribution it encounters in deployment.
A second interesting direction is a broader and tighter integration with complementary efficiency techniques. Our agentic optimization study shows that workload-aware caching, plan-level caching, and parallel execution are complementary and mutually reinforcing. However, these techniques currently operate independently. Jointly optimizing across caching, scheduling, and execution ordering could be interesting. For instance, a scheduling-aware cache that prioritizes retaining entries for agents scheduled to run soon, similar in spirit to KVFlow's approach at the KV cache level, could yield compounding efficiency gains beyond what is achievable by combining them independently.
Another interesting future direction is cache-aware planning, where the planning agent is informed of the current cache state when generating execution plans. Currently, planning and caching operate as independent stages, where the planner takes queries and generates DAG represented plans without knowledge of what is already cached. A planner aware of cached entries could decompose queries in ways that explicitly maximize reuse, representing a tighter and potentially more powerful integration of planning and caching in multi-agent systems.
\section{Conclusion}
In conclusion, we presented a workload-aware cache eviction policy for multi-agent systems that exploits three signals unique to DAG-structured agentic workloads, namely recomputation cost, dependency count, and agent invocation frequency. Unlike traditional eviction policies that treat all cached entries uniformly based on access history, our approach preserves entries that are expensive to recompute, structurally central to the DAG, and aligned with the current workload focus. Evaluated across three diverse benchmarks spanning different reuse regimes, our policy consistently outperforms all finite-capacity baselines in latency and throughput while maintaining accuracy, and approaches the performance of an unbounded cache despite operating under a fixed memory budget. 

Our results demonstrate that the quality of eviction decisions matters as much as cache hit rate in heterogeneous agentic workloads. By selectively retaining expensive and structurally important nodes, our policy converts a moderate hit rate advantage into large latency savings, a property that cost-blind policies cannot replicate. Our analysis of complementary techniques further shows that workload-aware content caching, plan-level caching, and parallel execution target distinct bottlenecks in multi-agent pipelines and are very effective in combination, with the right mix depending on workload structure. 

As multi-agent systems grow in complexity and scale, efficient resource utilization at every level of the execution stack becomes more and more important. We hope this work motivates further exploration of workload-aware optimization techniques for agentic systems, and highlights the value of using signals that are uniquely available in DAG-based multi-agent execution environments.

\bibliographystyle{IEEEtran}
\bibliography{references}

\end{document}